\documentclass{article} 
\usepackage[final]{colm2026_conference}

\usepackage{graphicx}
\usepackage[normalem]{ulem}

\usepackage{microtype}
\usepackage{hyperref}
\usepackage{url}
\usepackage{booktabs}
\usepackage{amsmath}
\usepackage{comment}
\usepackage[linesnumbered,ruled,vlined]{algorithm2e}

\usepackage{lineno}

\definecolor{darkblue}{rgb}{0, 0, 0.5}
\hypersetup{colorlinks=true, citecolor=darkblue, linkcolor=darkblue, urlcolor=darkblue}

\def\ours{{SCOPE}}
\title{SCOPE: A Generative Approach for LLM Prompt Compression}

\author{Tinghui Zhang \\
  University of Florida \\
  \texttt{tinghui.zhang@ufl.edu} \\\And
  Yifan Wang \thanks{Corresponding author.} \\
  University of Hawaii - Manoa \\
  \texttt{yifanw@hawaii.edu} \\\And
    Daisy Zhe Wang \\
 University of Florida \\
  \texttt{daisyw@cise.ufl.edu} \\
}

\begin{document}

\ifcolmsubmission
\linenumbers
\fi

\maketitle

\begin{abstract}

A big issue in modern LLM applications is they tend to feed long context to LLM, which results in high inference cost and latency, and may exceed the context limit.  
Prompt compression addresses this issue by reducing the length of input context with minimum loss of generation quality, i.e, the goal of prompt compression is to shorten the LLM input while maintaining a high generation quality.  
To overcome these limitations, we propose \ours{}, a training-free generative prompt compression framework based on chunk-level rewriting. Unlike the existing token removal methods, our method centers at a chunking-and-summarization mechanism. Specifically, \ours{} splits a prompt into semantically coherent chunks and rewrites the chunks to be more concise. Then the chunks are reconstructed into a meaningful prompt. Additionally, we design several optimization techniques for \ours, 
effectively preserving critical information and text coherence in compression, as well as providing finer-grained control of the compression ratio.  
We conduct extensive evaluation on typical LLM applications like question-answering and summarization.
Results show that \ours{} consistently outperforms the evaluated selective compression baselines across most settings, with particularly strong gains at high compression ratios.

\end{abstract}


\section{Introduction}
\vspace{-10pt}

With the increasing length of input context in Large Language Models (LLMs) applications, users/developers have to face significantly inference cost and latency increase, as well as risks exceeding model context limits. 
To address these issues, prompt compression methods have been developed to shorten prompt while retaining necessary information. The state-of-the-art methods are usually based on \textit{selective compression}, which focus on identifying and retaining the most important tokens of the original prompt, while removing less critical tokens. Although such approaches (called ``selective'' or ``token removal'' methods in this paper) reduce prompt length, they have notable drawbacks:
\textbf{(1) Breaking contextual integrity:} By removing tokens of the prompt, selective methods disrupt the original context, leading to a loss of coherence and a potentially lost conjunction part of different chunks.
\textbf{(2) Limiting context scope:} Selective methods use token as the processing unit, instead of considering at a higher level of the context, which limits their ability to understand the broader semantics.  
\textbf{(3) Losing key information:} Removal of tokens  potentially discards key information, which never be compensated after the compression.

To address the problems, we propose \ours, a generative prompt compression method based on prompt rewrite with generative model. \footnote{
Code is available at \url{https://github.com/tzhan024/scope-prompt-compression}.}
Specifically, \ours\ rewrites the original prompt to be more concise, while preserving the overall meaning and critical information. 
To balance higher-level semantics and lower-level details, the rewrite is conducted on chunk-level (i.e., chunk by chunk) rather than the whole context. 
\ours{} deploys a summarization model for the rewrite. And because the goal of prompt compression is to reduce LLM cost, it makes no sense to use another LLM for the summarization in our method, therefore a smaller summarization model (e.g., BART ~(\cite{lewis2019bart})) is chosen.

Our method involves the following steps: (1) \ours\ first breaks down the original prompt into semantically coherent chunks (size not fixed), ensuring each chunk covers a complete and meaningful piece of information. (2) Each chunk is then evaluated for its relevance to the original prompt, and sorted by the relevance ascendingly, then the \textit{less relevant} chunks are prioritized in compression. This is to squeeze the low-relevance chunks such that more space will be left for keeping highly relevant chunk content. 
(3) Furthermore, based on relevance and desired compression ratio, \ours\ calculates a specific (and usually different) compression ratio for each chunk. Such a dynamic compression ratio strategy guarantees the compression quality and robustness in finer grind. 
(4) Then each chunk is summarized in the sorted order with a summarization language model. Finally, the compressed (summarized) chunks are reassembled in their original order to form a coherent and shorter prompt.

Comparing to the state-of-the-art works like LLMLingua-2~(\cite{pan2024llmlingua}), \ours\ requires no additional training and performs compression using lightweight pretrained embedding and summarization models, without invoking large LLMs during compression. This distinguishes SCOPE from prior generative approaches that rely on fine-tuning or reinforcement learning (e.g., RECOMP, PRCA), making it suitable as a plug-and-play module for existing LLM systems. Instead, all models (mainly the embedding and summarization models) are public pre-trained small language models, making \ours\ resource efficient and easy to use.      
In terms of end-to-end running speed, the most time-consuming step in \ours\ is the summarization, which is not an issue. Our evaluation shows that a small summarization model (with fast inference speed) like BART~(\cite{lewis2019bart}) is enough to achieve significantly high compression quality.    
SCOPE is a training-free generative prompt compression framework that operates at the chunk level without requiring additional model training.
We evaluate our method on long document summarization and question-answering (QA) tasks, two common scenarios that require long input context. Our results show that \ours\ outperforms all the baselines in most cases for different downstream LLMs, effectively preserving semantic integrity and successfully maintaining inference quality after compression.

Our main contributions are as follows:

(1) We propose \ours\, a training-free generative prompt compression framework that performs semantic chunking and chunk-level rewriting using lightweight pretrained models, based on a novel chunking-and-summarization mechanism, which effectively addresses contextual integrity break and key information loss, and achieves significantly high performance. 

(2) We design several novel optimizations to the compression, including semantic chunking, outlier chunk handling, dynamic compression ratio, sorting-based compression priority and keyword maintaining. These techniques effectively improve the compression quality. 

(3) We conduct extensive evaluation for the end-to-end performance and critical optimizations of our method. The evaluation results affirm the effectiveness of our design and provide guidance for better using \ours.

In the rest of this paper, Section 2 introduces prior works related to prompt compression. Section 3 introduces the technical details of our method. Section 4 is the experiment section, reporting and analyzing the results of end-to-end evaluation and ablation studies.

\begin{figure*}[t]
  \centering
  \includegraphics[draft=false, width=1\textwidth]{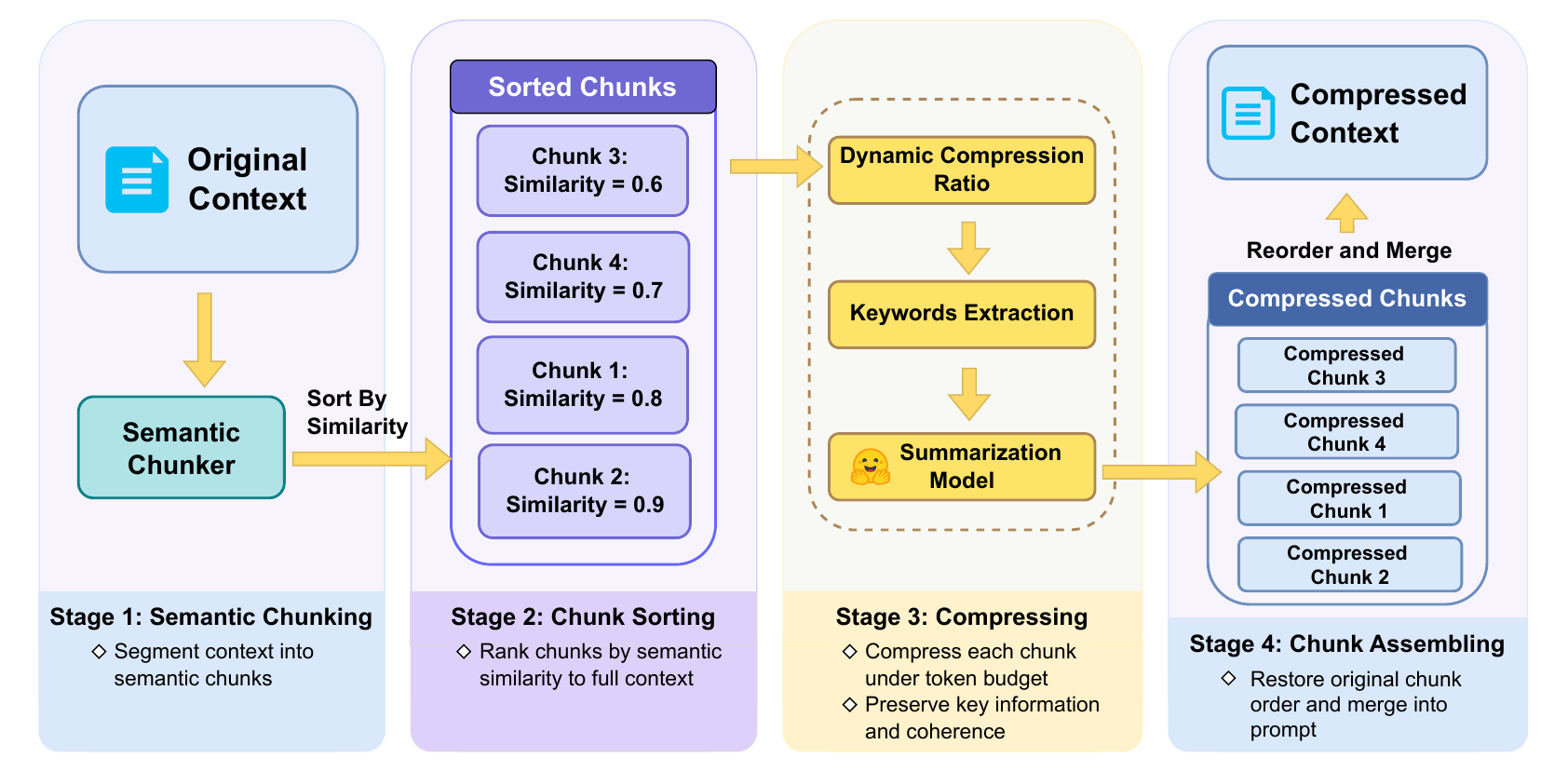}
  \caption{The workflow of SCOPE}
  \label{fig:pipeline-scope}
\end{figure*}

\section{Related Work} 
\vspace{-10pt}

Efficient handling of prompt lengths has become an important research area in large language models (LLMs), mainly to reduce computing costs and generation latencies. The LLMLingua series includes several methods to solve these issues.
Selective-Context (\cite{li2023unlocking}) is the earliest method, it selects tokens based on information theory but processes text in a single direction, which may lost important content.
LLMLingua (\cite{jiang2023llmlingua}), LongLLMLingua (\cite{jiang2023longllmlingua}), and LLMLingua-2 (\cite{pan2024llmlingua}) are selective prompt compression methods that estimate token importance to remove redundant content, but by largely treating the prompt as a single sequence, they may drop or overlook widely separated yet critical tokens.
Besides the LLMLingua methods, other approaches have also been developed for compressing prompts effectively. 
PartPrompt (\cite{mao2024parse}) performs structure-aware prompt compression by leveraging parse trees to estimate both local and global token importance and removing low-importance tokens accordingly.
COMPACT (\cite{yoon2024compact}) focuses on multi-document QA task, adaptively compressing documents based on query relevance and stopping once sufficient information has been gathered.
Style-Compress (\cite{pu2024style}) studies how different compression styles, affect downstream task performance and iteratively selects the most effective compressed prompt for each task.
Mean-Pooling Compression (\cite{feldman2025simple}) achieves fast, parameter-free compression by applying non-overlapping mean pooling over LLM hidden states, but its reliance on TPUs limits practicality in real-world settings.

In addition to selective (token-removal) approaches, several recent works
have explored generative or abstractive prompt compression.
RECOMP (\cite{xu2023recomp}) formulates compression as a learned rewriting
process, while PRCA (\cite{yang2023prca}) leverages reinforcement learning
for adaptive context selection. Prompt-SAW (\cite{ali2024prompt}) explores
abstractive compression through structured rewriting.
Compared to these approaches, SCOPE focuses on a training-free setting,
using lightweight summarization models and chunk-level control to achieve
efficient and flexible compression.

\section{Methodology}
\vspace{-10pt}

In this section, we introduce the overall workflow and specific optimization techniques in \ours.  
In Figure~\ref{fig:pipeline-scope}, we illustrate the four-stage pipeline of \ours.
\textbf{Stage 1} segments the original context (blue box) into semantic chunks using a semantic chunker (green box).
\textbf{Stage 2} ranks the chunks by their semantic similarity to the full context.
\textbf{Stage 3} applies dynamic chunk-level compression, where the compression ratio is adaptively determined based on similarity, chunk length, and token budget constraints, while preserving key entities and facts via keyword extraction and a generative summarization model.
\textbf{Stage 4} reorders and merges the compressed chunks to form the final compressed context.
To make it simple, we use "Line X" in this paragraph to denote the corresponding line in Algorithm~\ref{alg:overall}.  

In end-to-end workflow, the original context is first fed into the semantic chunker, which splits the entire context into chunks and computes similarity/relevance scores for each chunk to the full context (Line 1-2). Subsequently, keywords are extracted and maintained from each chunk (Line 3) for later postprocessing. The chunks are then sorted based on their similarity scores to the full context, ascendingly (Line 4), i.e., placing the less relevant chunks at the first. 
In the next step, a distinct compression ratio is calculated for each chunk individually (Line 5-7), followed by summarizing the chunk using the summarization model (Line 8-9). The resulting compressed chunks are then merged in original order to generate the final compressed context (Line 10).

Following prior work (\cite{liu2024retrievalattention}) on attention sink effects in transformers,
we preserve head and tail tokens (128 and 256 respectively),
as boundary tokens are known to receive disproportionately high attention.
Based on that, we propose two variants of \ours: (1) $\text{SCOPE}_D$, a \emph{head-and-tail preserving} variant that retains the initial and local window tokens, and (2) $\text{SCOPE}_N$, a \emph{no-preserving} variant that applies compression to all tokens.

\begin{algorithm}[t]
\small
\caption{End-to-End Prompt Compression}
\label{alg:overall}
\KwIn{text string $\mathit{essay}$, compression ratio $\rho$}
\KwOut{compressed text string}
\BlankLine
$E \gets \textsf{get\_full\_context\_embedding}(\mathit{essay})$\;
$\mathit{chunks} \gets \textsf{recursive\_semantic\_chunking}(\mathit{essay}, E)$\;

  ${keywords} \gets \textsf{extract\_important\_entities}(\mathit{chunks})$\;

$\mathit{chunks} \gets \textsf{ascending\_sort\_by\_sim}(\mathit{chunks, \mathit{essay}})$\;
$total \gets \textsf{count\_tokens}(\mathit{essay})$\;
$target \gets \lfloor total / \rho\rfloor$\;

$\mathit{ratios} \gets \textsf{get\_compression\_ratios}(\mathit{chunks}, \mathit{total}, \mathit{target})$\;
\BlankLine
\ForEach{$c \in \mathit{chunks}$}{

    compress $c$ based on $ratios[c]$ and $keywords[c]$ 
}
Concatenate all chunks in their original order in essay;

\BlankLine
\Return{$merged\ chunks$}\;
\end{algorithm}

\subsection{Semantic Chunking}
 
The chunking is conducted over three levels: section, paragraph, and sentence.  
It begins by recursively dividing long contexts into sections and/or paragraphs. If a paragraph has exceeded length, it will be split into finer chunks (of sentences), which is discussed in next section.   
The overall semantic chunking algorithm is described as follows:

\noindent\textbf{Step 1}: The whole context is first split by section. For each section, if its length falls within a predefined token limit (e.g., up to 800 tokens in our evaluation), it is treated as a single chunk; otherwise it will be further split by paragraph naturally, and each paragraph will be processed as in Step 2. 

\noindent\textbf{Step 2}: If a paragraph's length falls within the same limit, it is treated as one chunk. Otherwise it will be further split into chunks of sentences, where each chunk includes multiple semantically coherent sentences. The semantic coherence is measured by sentence embedding. More details are introduced in Section~\ref{sec:outlier}.  

\noindent\textbf{Step 3}: Finally, \ours\ merges adjacent tiny chunks into one chunk (by a length threshold) to preserve coherence and maintain a relatively stable chunk size.

Such a semantic chunking strategy splits the long context into manageable pieces while effectively preserving the semantic coherence within each piece of text. We evaluate the impact of semantic chunking in Section~\ref{sec:eval-chunk}.

\subsection{Paragraph Chunking}
 
\label{sec:outlier}
The chunking algorithm decides the proper split points to segment the text, aiming to maximize internal semantic consistency within each segment while balancing chunk size. The split points are selected based on semantic similarity between sentences.  
Specifically, in a paragraph, our chunking method first estimates how strongly each pair of adjacent sentences are related (by similarity between sentence embeddings), based on which it selects the initial split points where adjacent sentence similarity is less than a threshold. This results in the initial chunks. Then, the method grows each chunk up (with a upper size limit), until the end of the current chunk reaches a point that maximizes chunk-internal similarity and minimizes cross-chunk similarity. 
The internal similarity of a chunk is the average of adjacent sentence similarities in the chunk, while cross-chunk similarity is the similarity between the two sentences around the chunk end.  
If the remaining text is too short or a chunk exceeds the limit, it adjusts the split point or merges outlier chunks so that all final segments are well-balanced and semantically coherent.
More details are presented in Algorithm~\ref{alg:chunk-long} (Appendix).

\subsection{Compressing Chunks}
 
\subsubsection{Dynamic compression ratio}
 
Once the semantic chunking process is complete, we calculate the cosine
similarity (reflecting the relevance) between each chunk and the original
context. When the context exceeds the embedding model's input limit, we
encode it in smaller segments and average their embeddings to obtain an
aggregated full-context representation. Chunk relevance is then calculated
using cosine similarity with this representation. Using these similarity
measures, we derive a formula to determine the unique compression ratio for
each chunk. The principle is having a higher compression ratio for less relevant chunks while lowering compression ratio for more relevant chunks. In short, we squeeze the irrelevant chunks to leave space for keeping relevant chunks as much as original.   
Below are the details for the ratio calculation:

First, compute the number of tokens to be compressed out:
    \[
      L_{\mathrm{total}} = \sum_{i} L_{i},
      \qquad
      \Delta = L_{\mathrm{total}} - L_{\mathrm{total}}/\rho.
    \]
Where \(L_{\mathrm{total}}\) is the total token of the input context, \(L_{i}\) is the total token of each chunk, $\rho$ is the target overall compression ratio. So \(L_{\mathrm{total}}/\rho\) is the expected number of tokens after compression, then \(\Delta\) is the number of tokens to be reduced.

Second, define each chunk’s compressibility weight as 
    \[
      w_{i} = L_{i}\,(1 - S_{i}),
    \]
where \(S_{i}\) is the normalized similarity (relevance) score of chunk \(i\) (up to 1.0). This weight follows such a trend: (1) it decreases for higher relevance, and (2) it increases for longer chunk. We use it as the weight to determine the number of reduced tokens for each chunk, such that (1) relevant chunks (whose weight is small) tend to remain, and (2) long chunks (whose weight is large) tend to be compressed more significantly.      

Third, calculate difference of tokens after compression:
    \[
      \Delta_i 
      = \Delta \frac{w_i}{\sum_{i} w_{i}}
      = \Delta\frac{L_{i}(1 - S_{i})}{\sum_{i} L_{i}(1 - S_{i})}.
    \]
where $\Delta_i$ is the number of tokens to be reduced for chunk $i$.

Finally, the post-compression length of chunk $i$ is:
    \[
      L'_{i} 
      = L_{i} - \Delta_i.
    \]

By such, we compress each chunk with a dynamic compression ratio based on the weight of chunk size and relevance, effectively preserving critical information, as well as squeezing enough to guarantee the overall ratio close to the target.

\subsubsection{Compressing and postprocessing}
 
To maintain the original meaning of most critical text, we sort the chunks based on their similarity (relevancy) scores to the original full context from low to high. Then we start the compression from lower relevance chunks to higher, with the dynamic compression ratio. 
In addition, the compression will early stop once the target number of reduced tokens ($\Delta$) is reached. As a result, chunks with higher relevance to the original context are more likely to be fully preserved.

The compression is conducted using a generative summarization model, guided explicitly by the calculated compression ratio. Specifically, \ours\ instructs the model (like BART) to generate a summary of the given chunk $i$, and requires the output summary length to be $L^\prime_i$. 

Additionally, during compression, we extract and preserve critical keywords or factors within each chunk, crucial for maintaining effectiveness in downstream question-answering tasks.  The summarization model is always instructed to pay additional attention to these keywords when rewriting the text. Furthermore, if the compression result is overly truncated (i.e., too short), the result will be substituted with these preserved keywords, which acts as a postprocessing to maintain the key information.  
We extract keywords using NER,
which are enforced during summarization to preserve critical information.
Extremely short segments are merged with adjacent chunks to maintain coherence.
To quantify how frequently this safeguard is triggered, we analyze the first
200 GovReport examples under 5$\times$, 3$\times$, and 2$\times$ compression.
No excessively short chunks or keyword replacement cases are observed across
more than 7,000 generated chunks, indicating that such cases are rare in the
evaluated settings.

\begin{table*}[t]
\vspace{-6pt}
\centering
\scriptsize
\setlength{\tabcolsep}{2.5pt}
\caption{Evaluation on the Arxiv-Summarization dataset at different compression ratios on GPT-4o-mini and Qwen-2.5.}
\resizebox{\textwidth}{!}{
\begin{tabular}{l|c|c|c|c|c|c|c|c|c|c|c|c|r}
\toprule
\multicolumn{14}{c}{\textbf{Arxiv-Summarization (GPT-4o-mini / Qwen-2.5)}} \\
\midrule
Method &
\multicolumn{2}{c|}{BLEU (\%)} &
\multicolumn{2}{c|}{ROUGE-1 (\%)} &
\multicolumn{2}{c|}{ROUGE-2 (\%)} &
\multicolumn{2}{c|}{ROUGE-L (\%)} &
\multicolumn{2}{c|}{ROUGE-Lsum (\%)} &
\multicolumn{2}{c|}{BERTScore (\%)} &
$\rho$ \\
\cmidrule(lr){2-3}\cmidrule(lr){4-5}\cmidrule(lr){6-7}\cmidrule(lr){8-9}\cmidrule(lr){10-11}\cmidrule(lr){12-13}
& GPT & QWEN & GPT & QWEN & GPT & QWEN & GPT & QWEN & GPT & QWEN & GPT & QWEN & \\
\midrule
Selective-Context & 35.95 & 33.05 & 33.44 & 33.96 & 7.16 & 7.49 & 18.55 & 18.12 & 26.61 & 26.32 & 79.75 & 79.29 & 5× \\
LLMLingua         & 35.78 & 33.77 & 30.75 & 29.19 & 4.48 & 3.62 & 17.48 & 16.10 & 25.01 & 23.15 & 79.21 & 78.51 & 5× \\
LongLLMLingua     & 36.23 & 34.55 & 32.58 & 31.07 & 5.76 & 4.81 & 18.17 & 16.89 & 26.12 & 24.40 & 79.71 & 79.07 & 5× \\
PartPrompt        & 36.20 & 32.85 & 33.10 & 33.82 & 6.25 & 7.20 & 18.37 & 18.19 & 26.52 & 26.50 & 79.78 & 79.46 & 5× \\
LLMLingua-2       & 36.57 & 35.62 & 35.53 & 34.11 & 8.04 & 7.08 & 19.14 & 18.00 & 27.81 & 26.12 & 80.55 & 80.12 & 5× \\
$\text{SCOPE}_N$  & \textbf{38.09} & \textbf{37.02} & 37.75 & 36.26 & 10.31 & 9.15 & \textbf{20.16} & 19.12 & \textbf{29.12} & 27.59 & \textbf{80.93} & 80.52 & 5× \\
$\text{SCOPE}_D$  & 37.65 & 36.72 & \textbf{37.80} & \textbf{38.33} & \textbf{10.36} & \textbf{11.08} & 20.15 & \textbf{20.34} & 29.09 & \textbf{29.24} & 80.80 & \textbf{80.78} & 5× \\
\cmidrule(lr){1-1}\cmidrule(lr){2-14}
Selective-Context & 36.30 & 34.02 & 35.26 & 35.83 & 8.62 & 9.07 & 19.22 & 18.99 & 27.68 & 27.49 & 80.10 & 79.80 & 3× \\
LLMLingua         & 35.69 & 34.15 & 32.08 & 30.44 & 5.30 & 4.32 & 17.85 & 16.59 & 25.73 & 23.91 & 79.49 & 78.94 & 3× \\
LongLLMLingua     & 36.66 & 35.32 & 34.00 & 32.45 & 6.81 & 5.79 & 18.67 & 17.43 & 26.94 & 25.17 & 80.11 & 79.56 & 3× \\
PartPrompt        & 37.07 & 34.55 & 35.89 & 36.71 & 8.44 & 9.46 & 19.55 & 19.56 & 28.20 & 28.34 & 80.46 & 80.22 & 3× \\
LLMLingua-2       & 36.81 & 36.00 & 37.08 & 35.42 & 9.32 & 8.22 & 19.74 & 18.60 & 28.75 & 27.05 & 80.73 & 80.33 & 3× \\
$\text{SCOPE}_N$  & \textbf{38.26} & \textbf{37.24} & 38.17 & 36.59 & 10.55 & 9.36 & 20.24 & 19.25 & 29.40 & 27.81 & \textbf{80.98} & 80.59 & 3× \\
$\text{SCOPE}_D$  & 37.87 & 37.00 & \textbf{38.26} & \textbf{38.79} & \textbf{10.66} & \textbf{11.44} & \textbf{20.42} & \textbf{20.66} & \textbf{29.41} & \textbf{29.56} & 80.93 & \textbf{80.93} & 3× \\
\cmidrule(lr){1-1}\cmidrule(lr){2-14}
Selective-Context & 36.87 & 35.12 & 36.85 & 37.41 & 9.76 & 10.49 & 19.80 & 19.93 & 28.61 & 28.63 & 80.46 & 80.32 & 2× \\
LLMLingua         & 36.02 & 34.77 & 33.97 & 32.34 & 6.77 & 5.75 & 18.44 & 17.24 & 26.80 & 25.03 & 79.94 & 79.46 & 2× \\
LongLLMLingua     & 37.44 & 36.25 & 35.75 & 34.07 & 8.20 & 6.99 & 19.41 & 18.13 & 28.06 & 26.25 & 80.63 & 80.12 & 2× \\
PartPrompt        & 38.02 & 36.07 & 38.43 & 38.94 & 10.53 & 11.43 & 20.65 & 20.81 & \textbf{29.78} & 29.84 & \textbf{81.09} & 80.91 & 2× \\
LLMLingua-2       & 36.96 & 36.43 & 38.14 & 36.37 & 10.29 & 9.02 & 20.22 & 19.12 & 29.37 & 27.68 & 80.86 & 80.54 & 2× \\
$\text{SCOPE}_N$  & \textbf{38.14} & \textbf{37.35} & 38.66 & 37.07 & 11.04 & 9.68 & \textbf{20.67} & 19.62 & \textbf{29.78} & 28.20 & \textbf{81.09} & 80.71 & 2× \\
$\text{SCOPE}_D$  & 37.78 & 37.32 & \textbf{38.74} & \textbf{39.47} & \textbf{11.09} & \textbf{12.12} & 20.63 & \textbf{21.08} & 29.77 & \textbf{30.12} & 81.02 & \textbf{81.10} & 2× \\
\cmidrule(lr){1-1}\cmidrule(lr){2-14}
Original          & 38.69 & 37.98 & 40.49 & 41.23 & 12.63 & 13.59 & 21.85 & 22.43 & 31.12 & 31.51 & 81.50 & 81.49 & 1× \\
\bottomrule
\end{tabular}
}
\label{tab:arxiv_summarization_gpt_qwen_nested}
\vspace{-15pt}
\end{table*}

\begin{table*}[t]
\vspace{-6pt}
\centering
\scriptsize
\setlength{\tabcolsep}{2.5pt}
\caption{Evaluation on the GovReport-Summarization dataset at different compression ratios on GPT-4o-mini and Qwen-2.5.}
\resizebox{\textwidth}{!}{
\begin{tabular}{l|c|c|c|c|c|c|c|c|c|c|c|c|r}
\toprule
\multicolumn{14}{c}{\textbf{GovReport-Summarization (GPT-4o-mini / Qwen-2.5)}} \\
\midrule
Method &
\multicolumn{2}{c|}{BLEU (\%)} &
\multicolumn{2}{c|}{ROUGE-1 (\%)} &
\multicolumn{2}{c|}{ROUGE-2 (\%)} &
\multicolumn{2}{c|}{ROUGE-L (\%)} &
\multicolumn{2}{c|}{ROUGE-Lsum (\%)} &
\multicolumn{2}{c|}{BERTScore (\%)} &
$\rho$  \\
\cmidrule(lr){2-3}\cmidrule(lr){4-5}\cmidrule(lr){6-7}\cmidrule(lr){8-9}\cmidrule(lr){10-11}\cmidrule(lr){12-13}
& GPT & QWEN & GPT & QWEN & GPT & QWEN & GPT & QWEN & GPT & QWEN & GPT & QWEN & \\
\midrule
Selective-Context & 58.38 & 56.97 & 38.87 & 26.16 & 10.14 & 8.05 & 17.44 & 13.85 & 22.20 & 16.92 & 81.49 & 80.69 & 5× \\
LLMLingua         & 52.63 & 53.73 & 34.62 & 25.12 & 6.22 & 4.60 & 15.80 & 12.72 & 19.83 & 15.09 & 79.88 & 79.74 & 5× \\
LongLLMLingua     & 55.14 & 56.62 & 36.68 & 25.97 & 8.05 & 6.03 & 16.61 & 13.22 & 20.95 & 15.54 & 80.69 & 80.45 & 5× \\
PartPrompt        & 56.30 & 55.41 & 38.28 & 26.78 & 9.03 & 7.45 & 17.07 & 13.94 & 21.76 & 17.29 & 80.94 & 79.86 & 5× \\
LLMLingua-2       & 58.29 & 60.34 & 40.46 & 25.10 & 11.16 & 7.56 & 18.15 & 13.36 & 23.10 & 15.14 & 81.28 & 80.96 & 5× \\
$\text{SCOPE}_N$  & 60.52 & 62.72 & 43.59 & 26.52 & 14.00 & 9.18 & 19.20 & 14.29 & 24.76 & 16.21 & 81.86 & 81.76 & 5× \\
$\text{SCOPE}_D$  & \textbf{60.84} & \textbf{63.24} & \textbf{43.73} & \textbf{27.14} & \textbf{14.17} & \textbf{10.52} & \textbf{19.56} & \textbf{14.97} & \textbf{24.89} & \textbf{17.61} & \textbf{82.01} & \textbf{82.31} & 5× \\
\cmidrule(lr){1-1}\cmidrule(lr){2-14}
Selective-Context & 59.38 & 58.38 & 40.58 & 26.53 & 11.58 & 8.97 & 18.20 & 14.22 & 23.19 & 17.31 & 81.80 & 81.18 & 3× \\
LLMLingua         & 53.45 & 55.22 & 36.25 & 25.43 & 7.22 & 5.29 & 16.29 & 12.87 & 20.69 & 15.23 & 80.35 & 80.21 & 3× \\
LongLLMLingua     & 56.26 & 58.15 & 38.67 & 26.12 & 9.55 & 6.85 & 17.35 & 13.47 & 22.14 & 15.66 & 81.14 & 80.89 & 3× \\
PartPrompt        & 58.66 & 57.65 & 40.44 & \textbf{27.42} & 11.02 & 8.63 & 18.01 & 14.37 & 23.09 & \textbf{17.75} & 81.58 & 80.60 & 3× \\
LLMLingua-2       & 59.93 & 61.64 & 42.46 & 25.77 & 12.80 & 8.40 & 19.06 & 13.74 & 24.28 & 15.59 & 81.81 & 81.36 & 3× \\
$\text{SCOPE}_N$  & 60.86 & 62.70 & 43.69 & 26.53 & 14.10 & 9.21 & 19.36 & 14.33 & 24.85 & 16.18 & 81.97 & 81.80 & 3× \\
$\text{SCOPE}_D$  & \textbf{61.04} & \textbf{63.19} & \textbf{43.73} & 27.13 & \textbf{14.26} & \textbf{10.53} & \textbf{19.62} & \textbf{14.99} & \textbf{24.97} & 17.71 & \textbf{82.11} & \textbf{82.25} & 3× \\
\cmidrule(lr){1-1}\cmidrule(lr){2-14}
Selective-Context & 60.15 & 59.19 & 41.72 & 26.67 & 12.67 & 9.27 & 18.83 & 14.37 & 23.92 & 17.38 & 82.03 & 81.49 & 2× \\
LLMLingua         & 54.61 & 56.89 & 38.31 & 25.69 & 8.69 & 6.20 & 16.99 & 13.09 & 21.87 & 15.35 & 80.85 & 80.65 & 2× \\
LongLLMLingua     & 58.56 & 60.46 & 40.77 & 26.67 & 11.54 & 8.05 & 18.29 & 13.98 & 23.35 & 16.12 & 81.71 & 81.43 & 2× \\
PartPrompt        & 60.32 & 59.66 & 41.79 & \textbf{27.76} & 12.61 & 9.66 & 18.80 & 14.77 & 23.94 & \textbf{18.01} & 82.04 & 81.35 & 2× \\
LLMLingua-2       & 60.92 & 62.02 & 43.60 & 26.06 & 14.06 & 8.86 & 19.71 & 14.02 & \textbf{24.99} & 15.91 & 82.10 & 81.56 & 2× \\
$\text{SCOPE}_N$  & 61.35 & 62.94 & \textbf{43.79} & 26.60 & \textbf{14.44} & 9.30 & 19.58 & 14.29 & 24.95 & 16.18 & 82.17 & 81.87 & 2× \\
$\text{SCOPE}_D$  & \textbf{61.65} & \textbf{63.73} & 43.75 & 27.38 & 14.41 & \textbf{10.86} & \textbf{19.78} & \textbf{15.22} & 24.94 & 17.96 & \textbf{82.26} & \textbf{82.47} & 2× \\
\cmidrule(lr){1-1}\cmidrule(lr){2-14}
Original          & 62.46 & 64.56 & 43.78 & 27.97 & 14.98 & 11.57 & 20.25 & 15.66 & 25.29 & 18.23 & 82.46 & 82.74 & 1× \\
\bottomrule
\end{tabular}
}
\label{tab:govreport_summarization_gpt_qwen}
\vspace{-6pt}
\end{table*}

\subsection{Reconstruction of Compressed Context}
 
After individual chunk compression, compressed chunks are reordered and merged according to their original sequence within the initial context. Additionally, original separators and paragraph boundaries are reinstated to maintain structural and narrative coherence. Then \ours\ outputs the merged context as final result.

\section{Experiment}
\vspace{-10pt}

All experiments are evaluated on a machine with one NVIDIA A100 80GB Tensor Core GPU, 3 AMD Rome cores CPU and 64GB RAM.
In this paper, we use $\rho$ to denote the compression ratio, which is calculated as uncompressed prompt length over length after compression. 

In our implementation, the semantic chunking threshold $\tau$ is set to 0.5,
and chunk sizes are constrained between 400 and 700 tokens.
The summarization model can be vary (e.g., bart-large-cnn).
Keyword extraction is implemented using the HuggingFace Transformers NER
pipeline with \texttt{dbmdz/bert-large-cased-finetuned-conll03-english},
with numerical and date expressions additionally preserved through
rule-based matching.

\subsection{Experiment Setting}
 
\noindent\underline{\textbf{(1) Datasets}}
For most datasets, we use the test sets for evaluation. For Trivia-QA, we use the training set since the test set does not provide ground-truth answers. Due to time and resource limit, we randomly sample 2,000 records (i.e., 2000 prompts to be compressed) from each test set. To verify that the sampled evaluation is representative, we additionally evaluate all methods on the full Arxiv-Summarization test set of 6,440 examples at 5$\times$ compression. The relative ranking of the evaluated methods remains unchanged; detailed results are reported in Appendix~\ref{sec:full_arxiv}. If a set includes less than 2000 rows, we use it all.   
For the summarization task, we utilize three datasets: \textit{Arxiv-Summarization}~(\cite{cohan2018discourse}) for arXiv paper summarization, \textit{Pubmed-Summarizaiton}~(\cite{cohan2018discourse}) for biomedical paper summarization, and \textit{GovReport-Summarization}~(\cite{huang2021efficient}) for long government report summarization.  
In this task, the given paper or report is first compressed by a compression method, 
then the compressed context is input to downstream LLM to generate a summary, which is compared against the ground truth summaries. 
For the question answering task, we employ two datasets: \textit{Trivia-QA} (\cite{joshi2017triviaqa}) for reading-comprehension and \textit{MultiFieldQA-en}~(\cite{bai2023longbench}) from LongBench for long document QA. 
In this task, context is first compressed by a compression method, then fed to downstream LLM to generate answers, which are subsequently compared to the ground truth answers. More details are in Appendix.

\setlength{\tabcolsep}{1.5pt}   
\renewcommand{\arraystretch}{0.95}
\begin{table}[t]
\vspace{-6pt}
\centering
\scriptsize

\begin{minipage}[t]{0.55\linewidth}
\centering
\caption{Evaluation for QA on Trivia-QA and Longbench Multifield-QA-en datasets on both GPT-4o-mini (by default) and Qwen-2.5 ("W/T Qwen") models.}
\scalebox{1.03}{
\begin{tabular}{l|ccc|r}
\toprule

\multicolumn{5}{c}{\textbf{Question Answering Results}} \\
\midrule
Methods           & Trivia-QA  & Multifield-QA &Trivia-QA W/T Qwen & $\rho$ \\
\cmidrule(lr){1-1}\cmidrule(lr){2-4}\cmidrule(lr){5-5}
Selective-Context & 82.31   & 25.82     & 66.99     & 5× \\
LLMLingua         & 80.05   & 23.35     & 58.02     & 5× \\
LongLLMLingua     & 80.79   & 27.03     & 60.04     & 5× \\
PartPrompt         & 82.61   & 26.82     & 63.95     & 5× \\
LLMLingua-2       & 83.58   & 36.23     & 72.68     & 5× \\
$\text{SCOPE}_N$      & 85.31  & 41.39    & 77.85          & 5× \\
$\text{SCOPE}_D$      & \textbf{85.66}  & \textbf{43.54}    & \textbf{78.36}          & 5× \\
\cmidrule(lr){1-1}\cmidrule(lr){2-4}\cmidrule(lr){5-5}
Selective-Context & 83.85   & 32.81     & 71.52     & 3× \\
LLMLingua         & 80.63   & 25.23     & 59.85     & 3× \\
LongLLMLingua     & 81.35   & 30.96     & 62.65     & 3× \\
PartPrompt         & 83.04   & 33.62     & 68.19     & 3× \\
LLMLingua-2       & 85.09   & 40.32     & 76.89     & 3× \\
$\text{SCOPE}_N$      & 85.57  & 42.22    & 78.18          & 3× \\
$\text{SCOPE}_D$      & \textbf{85.81}  & \textbf{46.09}    & \textbf{78.82}          & 3× \\
\cmidrule(lr){1-1}\cmidrule(lr){2-4}\cmidrule(lr){5-5}
Selective-Context & 85.55   & 40.24     & 76.22     & 2× \\
LLMLingua         & 81.58   & 30.16     & 62.15     & 2× \\
LongLLMLingua     & 84.40   & 35.10     & 70.58     & 2× \\
PartPrompt         & 84.70   & 40.97     & 73.53     & 2× \\
LLMLingua-2       & 85.84   & 46.96     & 79.16    & 2× \\
$\text{SCOPE}_N$      & 86.07  & \textbf{47.52}    & 79.56         & 2× \\
$\text{SCOPE}_D$      & \textbf{86.25}  & 47.15    & \textbf{80.57}         & 2× \\
\cmidrule(lr){1-1}\cmidrule(lr){2-4}\cmidrule(lr){5-5}
Original          & 86.08   & 54.20     & 81.99 & 1× \\

\bottomrule
\end{tabular}
}
\label{tab:qa_results}
\end{minipage}
\hfill
\begin{minipage}[t]{0.4\linewidth}
\centering
\caption{Ablation study for keyword extraction on Trivia-QA for different compression ratios with GPT-4o-mini.}
\scalebox{1}{
\begin{tabular}{l|c|r}
\toprule

\multicolumn{3}{c}{\textbf{Trivia-QA}} \\
\midrule
Method            & F1-Score (\%) & $\rho$ \\
\cmidrule(lr){1-1}\cmidrule(lr){2-2}\cmidrule(lr){3-3}
$\text{SCOPE}_N$   & \textbf{85.57}     & 3× \\
$\text{SCOPE}_N$ W/O Keyword Extraction      & 84.95     & 3× \\

\cmidrule(lr){1-1}\cmidrule(lr){2-2}\cmidrule(lr){3-3}
$\text{SCOPE}_N$     & \textbf{86.07}     & 2× \\
$\text{SCOPE}_N$ W/O Keyword Extraction      & 85.36     & 2× \\

\bottomrule
\end{tabular}
}
\label{tab:ablation_trivia_qa}
\end{minipage}

\vspace{-12pt}
\end{table}

\begin{table*}[t]
\vspace{-6pt}
\centering
\scriptsize
\setlength{\tabcolsep}{1pt}
\label{tab:ablation_chunking_ratio}
\caption{Ablation study for semantic chunking and dynamic compression ratio calculation, evaluated on Arxiv-Summarization and GovReport‐Summarization datasets at 2× compression ratio on GPT-4o-mini.}

\begin{tabular}{l|cccccc|r}
\toprule
\multicolumn{8}{c}{\textbf{Arxiv-Summarization}} \\
\midrule
Method            & BLEU (\%)   & ROUGE-1 (\%) & ROUGE-2 (\%) & ROUGE-L (\%) & ROUGE-Lsum (\%) & BERTScore (\%) & $\rho$ \\
\cmidrule(lr){1-1}\cmidrule(lr){2-7}\cmidrule(lr){8-8}
$\text{SCOPE}_N$      & \textbf{38.14} & 38.66 & 11.04 & \textbf{20.67} & \textbf{29.78} & \textbf{81.09} & 2× \\
$\text{SCOPE}_D$      & 37.78 & \textbf{38.74} & \textbf{11.09} & 20.63 & 29.77 & 81.02 & 2× \\
Fix Chunking \& compression ratio         & 37.00  & 37.98   & 10.53    & 20.10   & 29.24      & 80.75     & 2× \\

\toprule
\multicolumn{8}{c}{\textbf{GovReport-Summarization}} \\
\midrule
Method            & BLEU (\%)   & ROUGE-1 (\%) & ROUGE-2 (\%) & ROUGE-L (\%) & ROUGE-Lsum (\%) & BERTScore (\%) & $\rho$ \\
\cmidrule(lr){1-1}\cmidrule(lr){2-7}\cmidrule(lr){8-8}
$\text{SCOPE}_N$      & 61.35  & \textbf{43.79}   & \textbf{14.44}   & 19.58   & 24.95      & 82.17     & 2× \\
$\text{SCOPE}_D$      & \textbf{61.65}  & 43.75   & 14.41   & \textbf{19.78}   & 24.94      & \textbf{82.26}     & 2× \\
Fix Chunking \& compression ratio         & 60.73  & 43.13   & 13.87    & 19.28   & 24.65      & 81.93     & 2× \\

\bottomrule
\end{tabular}
\label{tab:ablation_chunking_ratio}
\vspace{-12pt}
\end{table*}

\noindent\underline{\textbf{(2) Baselines}}
We use five baseline methods: Selective-Context~(\cite{li2023unlocking}), LLMLingua~(\cite{jiang2023llmlingua}), LongLLMLingua~(\cite{jiang2023longllmlingua}), PartPrompt~(\cite{mao2024parse}), and LLMLingua-2~(\cite{pan2024llmlingua}). Without explicit notation, all baselines and our method are using GPT-4o-mini as the downstream LLM. In addition, we also test the performance of all methods on different LLMs, reported in Section~\ref{sec:different-llm}.
For our own methods, we use \ours{} to denote both variants (\ours$_N$ and \ours$_D$, introduced in Section 3) by default, and only distinguish them if necessary.

\noindent\underline{\textbf{(3) Metrics}}
For summarization tasks, we use \textit{ROUGE} (\cite{lin2004rouge}) (including ROUGE-1, ROUGE-2, ROUGE-L, and ROUGE-Lsum), \textit{BLEU}~(\cite{papineni2002bleu}), and \textit{BERTScore}~(\cite{zhang2019bertscore}) as metrics of summary quality. ROUGE and BLEU are both measuring the n-grams match between ground truth and generated result, while BERTScore uses pre-trained contextual embeddings from BERT models to evaluate the similarity between predicting and reference contexts.  For QA tasks, we evaluate the answer quality by \textit{F1-score}.
These metrics evaluate downstream task performance, which indirectly
reflects the effectiveness of prompt compression in preserving
task-relevant information.

\subsection{End-to-End Results on Summarization}
 
This section reports and analyzes the end-to-end performance of \ours, as well as baseline methods on the summarization task, under various compression ratios (5x, 3x, and 2x). 
The summaries generated from compressed contexts are evaluated against the dataset-provided reference summaries using BLEU, ROUGE, and BERTScore.
Tables~\ref{tab:arxiv_summarization_gpt_qwen_nested}, \ref{tab:govreport_summarization_gpt_qwen}, \ref{tab:pubmed_summarization} present summarization results across three datasets under various compression levels. We further evaluate SCOPE on the LongBench Multi-News dataset to examine its performance on multi-document summarization. SCOPE maintains the strongest overall performance among the evaluated methods across 2$\times$, 3$\times$, and 5$\times$ compression ratios; full results are provided in Appendix~\ref{sec:multinews}. The result and conclusion from \textit{Pubmed-Summarizaiton} are similar to other 2 datasets, so we place it (Table \ref{tab:pubmed_summarization}) in appendix due to space limit. We have following observations: 

(1) Our methods show the overall highest performance. They consistently achieve the highest scores in the vast majority of evaluation metrics across almost all summarization datasets. Specifically, our methods achieves up to 1.3x improvement of metric scores comparing to baseline methods, and outperforms the state-of-the-art methods in almost all cases. On Arxiv-Summarization dataset, SCOPE leads in all metrics under all compression ratios.  
On GovReport-Summarization, SCOPE is the best again under 5x and 3x compression. With 2x compression, SCOPE's overall performance remains dominant.  

(2) \ours\ has the most stable performance across different compression ratios. From 2 to 5x compression across different datasets, its score difference is always within 1\%, while all of the baseline methods show a larger score variation.
Such a stability underscores the robustness and adaptability of our method to varied domains.

(3) The advantage of \ours\ enlarges when compression ratio increases. 
Specifically, \ours\ excels at higher compression ratios (e.g., 5x). 
For example, in GovReport-Summarization at 5x compression with GPT as downstream LLM, \ours\ achieves a ROUGE-1 of 43.73 and ROUGE-2 of 14.17, while the closest competitor (LLMLingua-2) scores 40.46 and 11.16 respectively. This indicates the effectiveness of our method to preserve the most critical information even when context is significantly compressed.
Furthermore,  
\ours\ stably achieves a near-original performance (even under 5x compression), with scores remarkably close to the uncompressed ("Original") input.

\subsection{End-to-End Results on QA}
 
This section reports the results on QA datasets, at three compression ratios (5x, 3x, and 2x), presented in Table~\ref{tab:qa_results} 

(1) \ours\ consistently achieves the highest F1-Score in all cases. In Trivia-QA, \ours\ achieves up to 5\% higher F1-Score than the baselines, while achieving 20\% higher F1-Score in Multifield-QA-en, which shows the significant performance of our method and its outstanding capability to deliver most critical information to downstream LLM.
An interesting observation is that on Trivia-QA with GPT-4o-mini, \ours\ at 2x compression achieves an F1-score of 86.25, slightly higher than 86.08 with the original prompt. One possible explanation is that the rewriting process makes task-relevant information more salient to the downstream LLM.

(2) Similarly to summarization task, \ours\ presents the greatest stability across different compression ratios. As the compression ratio varies between 5x to 2x, the F1-Score of \ours\ remains more stable than the baselines. From 5x to 2x, \ours\ shows a variation less than 2\% and 4\% respectively on two datasets, while most baselines vibrate more than 2\% and 4\%. LLMLingua-2 even has a 10\% difference from 5x to 2x. 

(3) Same as summarization, the advantage of SCOPE is more obvious with higher compression ratio. The F1-Score gaps between \ours\ and the baselines are largest under 5× compression, particularly on MultiField-QA-en, where \ours\ achieves 20\% higher score than LLMLingua. 
This indicates that \ours\ is more effective than baselines at retaining question-relevant information when the context is heavily reduced.  

In summary, SCOPE maintains superior performance on both datasets under a wide range of compression ratios, highlighting its effectiveness for long-context question answering tasks.

\subsection{Evaluation on Different Models}
 
\label{sec:different-llm}
To evaluate the generalizability of our method, we evaluate some of the experiments with Qwen-2.5 (~\cite{yang2025qwen2}) as the downstream LLM. Table \ref{tab:arxiv_summarization_gpt_qwen_nested}, \ref{tab:govreport_summarization_gpt_qwen} and \ref{tab:qa_results} show the end-to-end results using Qwen-2.5 on different datasets and tasks given various compression ratios. 
We observe that SCOPE consistently achieves the highest scores in most cases of both summarization and QA tasks. 
And SCOPE also shows similar features to previous evaluation, like the highest stability given different metric scores and compression ratios, as well as reaching the biggest advantage over the baselines under 5x compression. All such observations indicate a strong generalizability and robustness of our method. 

We also examine the sensitivity of SCOPE to the compression model by replacing BART with Llama-3-8B-Instruct. The two compression backbones exhibit similar performance trends on GovReport and Multi-News across different compression ratios, suggesting that SCOPE is not limited to a particular summarization backbone. Detailed results are reported in Appendix~\ref{sec:compression_backbone}.

\subsection{Efficiency}
 
We compare the runtime of different compression methods across datasets and compression ratios (2x, 3x and 5x).  
Comparing to the baselines, SCOPE introduces moderate latency. For example, on GovReport dataset, across all three compression ratios, averagely SCOPE takes 4.20 seconds, PartPrompt takes 4.49 seconds, while LongLLMLingua takes 2.79 seconds; on Arxiv dataset, the three runtime numbers are 3.39s, 3.96s, and 2.15s. Given the significant quality improvement, such a slight latency overhead is acceptable, which ensures our method to be practical in real world.

\subsection{Ablation studies}
 
This section presents ablation experiment results for several key components of SCOPE: (1) semantic‐chunking + dynamic compression ratio, (2) keywords extraction, (3) the embedding models and summarization models. 

For the first, we evaluate \ours\ with semantic‐chunking + dynamic compression ratio VS. \ours$_N$ with fixed length chunking + fixed compression ratio, using Arxiv-Summarization and GovReport-Summarization datasets and 2x compression ratio. 
For the second, since keywords are more important to QA tasks, we evaluate it on Trivia-QA dataset and both 3x and 2x compression. Specifically, we compare \ours\ with VS without keyword extraction in QA task. 
For the last, we deploy different embedding models and summarization models in \ours\ and compare the end-to-end performance between them. The evaluations are both conducted on summarization task, and both results are reported in Table~\ref{tab:embedding_summ_horizontal_strict}, please see appendix for more details about this result.

\noindent\underline{\textbf{(1) Impact of chunking method}}
\label{sec:eval-chunk}
As in Table~\ref{tab:ablation_chunking_ratio}, on both summarization datasets, the full SCOPE pipeline with semantic chunking and dynamic ratio, outperforms that with fixed length chunking and fixed compression ratio in all cases. 
With the semantic chunking and dynamic ratio, the full \ours\ gets further performance gain, proving the importance of those optimizations.

\noindent\underline{\textbf{(2) Impact of keyword maintaining}}
In Table ~\ref{tab:ablation_trivia_qa}, SCOPE without keyword extraction on Trivia-QA leads to consistent F1-Score drop under both 3x and 2x compression. This indicates that explicitly highlighting question‐relevant terms is critical for LLM to generate accurate answer. 
By identifying keywords before compression and maintaining (adding back) them after compression, SCOPE effectively compensates for information loss. 

\noindent\underline{\textbf{(3) Embedding and summarization models}}

The result in Section~\ref{sec:appendix-addi-exp} and Table~\ref{tab:embedding_summ_horizontal_strict} shows SCOPE maintains a stable, high performance with different summarization and embedding models. Please see Appendix for more details.

\section{Conclusion}
\vspace{-10pt}

In this paper, we introduce SCOPE, a training-free generative prompt
compression framework that splits long contexts into semantically coherent
chunks and rewrites each chunk using an off-the-shelf summarization model. 
Evaluation shows \ours\ consistently outperforms token-removal approaches, 
especially under high compression ratio, demonstrating the effectiveness of \ours\ across the evaluated settings. We believe our work can effectively save the cost in LLM application, thereby contribute to the advance of AI community. 

\bibliography{colm2026_conference}
\bibliographystyle{colm2026_conference}

\appendix

\section{Appendix}

\label{sec:appendix}

\subsection{Details of Chunking Paragraphs}
\begin{algorithm}[H]
\small
\caption{Chunking Paragraph}
\label{alg:chunk-long}
\KwIn{text string $T$, chunk size bound $max\_token$, $min\_token$, similarity threshold $\tau$}
\KwOut{list of text chunks}
\BlankLine
Split $T$ into sentences $S$\;
Count tokens for each sentence: $C \gets [\,\text{countTokens}(s)\mid s\in S\,]$\;
Compute sentence embeddings $E$ and similarities $\text{sim}$ between adjacent sentences\;
Identify candidate split points: $\text{cand} \gets \{i+1 \mid \text{sim}[i] < \tau\}$\;
\BlankLine
Initialize $\text{bounds} \gets []$, $start \gets 0$\;
\While{$start < |S|$}{
    \If{remaining tokens $< min\_token$ and bounds not empty}{
        Merge or rebalance the last chunk if needed\;
        \textbf{break}
    }
    Expand chunk greedily up to $max\_token$ tokens from $start$\;
    Find candidate split points in $[start, end]$ with valid token count\;
    \eIf{such points exist}{
        Choose the split with the best semantic score
    }{
        Split at $end$
    }
    Adjust cut if needed to keep chunk $\leq max\_token$ tokens\;
    Append $(start, cut)$ to bounds\;
    $start \gets cut$
}
Build chunks by joining the sentences between each $(start, end)$ in bounds\;
\Return{list of text chunks}
\end{algorithm}

Algorithm ~\ref{alg:chunk-long} outlines the workflow of our paragraph-level chunking method. The primary components include token counting, embedding computation, identifying potential split points based on sentence similarity, and determining the start and end indices for each chunk.

Lines 1–3 initialize the algorithm by counting tokens in each sentence, encoding sentences to embeddings, and calculating adjacent sentence similarities. Lines 4 and 5 initialize variables for the split points of the candidate chunks and starting points of the iteration. 

Lines 6–18 iteratively decide how to segment the text, aiming to balance chunk size while maximizing internal semantic consistency within each segment. Line 8 checks if the remaining text is too short to form a valid chunk, triggering a merge or adjustment if needed. Lines 10 and 11 expand each segment up to the maximum token limit, and determine the split points for the potential chunks. Lines 12–17 select the most suitable split point based on semantic similarity, optimizing the segmentation for coherence.

Finally, the last 2 lines generate the finalized chunk list from determined chunk boundaries (the split points), returning coherent chunks as a list.

\subsection{Dataset Details}
\textbf{Arxiv-Summarization} (\cite{cohan2018discourse}) is a large-scale dataset for scientific paper summarization, which contains both section-level and document-level granularity. This dataset contains about 216k examples, it pairs full-text arXiv papers that are up to 745,000 characters, with their abstracts that are up to 92,700 characters in each row. All data is in English, and both section and document subset contains about 216k rows, which split about 203k rows in train set, 6440 rows in validation set, and 6440 rows in test set. 

\textbf{PubMed-Summarization} (\cite{cohan2018discourse}) is a biomedical summarization dataset containing pairs of article and abstract with about 133,215 rows in total. It pairs full-text PubMed articles of up to 826,000 characters with their abstracts of 174 to 4,340 characters in each row. All data is in English and the dataset is split into 119,924 rows in the train set, 6,633 rows in the validation set and 6,658 rows in the test set. 

\textbf{GovReport-Summarization} (\cite {huang2021efficient}) is a dataset for summarization of long government reports containing about 19,463 pairs of document and summary. Each row pairs a government report of 320 to 1,320,000 characters with its concise summary of 127 to 13,700 characters . All data is in English and the document subset is split into 17,517 rows in the train set, 973 rows in the validation set and 973 rows in the test set. 

\textbf{Trivia-QA} (\cite{joshi2017triviaqa}) is a reading-comprehension dataset containing triples of question, answer, and evidence. The subset used in our experiement \texttt{rc.wikipedia} contains about 77,600 rows. Each row provides a question authored by trivia enthusiasts, associate evidence documents and the ground-truth answer. \texttt{rc.wikipedia} subset is split into about 61,900 rows in the train set, 7,990 rows in the validation set and 7,700 rows in the test set. 

\textbf{MultiFieldQA-en} (\cite{bai2023longbench}) is a subset of the LongBench dataset, designed to evaluate large language models' (LLMs) ability to comprehend and answer questions based on long, single-document contexts across diverse domains. This subset contains 150 question-answering rows, each accompanied by a lengthy context document. The documents span various fields, including academic papers, legal documents, government reports, and technical manuals, reflecting real-world scenarios where in-depth understanding is required.

\begin{table*}[t]
\centering
\scriptsize
\setlength{\tabcolsep}{4pt}
\caption{Evaluation for Pubmed-Summarization dataset for different methods and compression ratio on GPT-4o-mini.}
\resizebox{\textwidth}{!}{
\begin{tabular}{l|cccccc|r}
\toprule
\multicolumn{8}{c}{\textbf{Pubmed-Summarization}} \\
\midrule
Method            & BLEU (\%)   & ROUGE-1 (\%) & ROUGE-2 (\%) & ROUGE-L (\%) & ROUGE-Lsum (\%) & BERTScore (\%) & $\rho$ \\
\cmidrule(lr){1-1}\cmidrule(lr){2-7}\cmidrule(lr){8-8}
Selective-Context & 43.92  & 34.67   &  8.30   & 18.37   & 27.89      & 81.74     & 5× \\
LLMLingua         & 43.27  & 29.97   &  5.72   & 16.32   & 24.55      & 80.97     & 5× \\
LongLLMLingua     & 44.66  & 32.61   &  7.44   & 17.48   & 26.39      & 81.71     & 5× \\
PartPrompt     & 43.17  & 32.56   &  6.52   & 17.28   & 26.46      & 81.04     & 5× \\
LLMLingua-2       & 45.02  & 35.86   &  8.60   & 18.74   & 28.61      & 81.91     & 5× \\
$\text{SCOPE}_N$      & \textbf{45.93}  & \textbf{38.03}   & 10.68   & 19.92   & \textbf{30.25}      & 82.26     & 5× \\
$\text{SCOPE}_D$      & 45.64  & 38.00   & \textbf{10.85}   & \textbf{19.99}   & 30.18      & \textbf{82.34}     & 5× \\
\cmidrule(lr){1-1}\cmidrule(lr){2-7}\cmidrule(lr){8-8}
Selective-Context & 44.76  & 36.56   &  9.69   & 19.30   & 29.24      & 82.22     & 3× \\
LLMLingua         & 43.38  & 31.58   &  6.40   & 16.93   & 25.58      & 81.23     & 3× \\
LongLLMLingua     & 45.27  & 34.22   &  8.31   & 18.22   & 27.59      & 82.03     & 3× \\
PartPrompt     & 44.90  & 35.60   &  8.71   & 18.75   & 28.67      & 81.92     & 3× \\
LLMLingua-2       & 45.43  & 37.53   & 10.02   & 19.62   & 29.77      & 82.31     & 3× \\
$\text{SCOPE}_N$      & 46.07  & 38.42   & 11.05   & 20.20   & \textbf{30.58}      & 82.45     & 3× \\
$\text{SCOPE}_D$      & \textbf{46.08}  & \textbf{38.46}   & \textbf{11.20}   & \textbf{20.24}   & 30.51      & \textbf{82.50}     & 3× \\
\cmidrule(lr){1-1}\cmidrule(lr){2-7}\cmidrule(lr){8-8}
Selective-Context & 45.80  & 38.15   &  10.88   & 20.12   & 30.34      & 82.64     & 2× \\
LLMLingua         & 43.89  & 33.48   &  7.47   & 17.77   & 26.88      & 81.63     & 2× \\
LongLLMLingua     & 46.12  & 36.10   &  9.38   & 19.11   & 29.02      & 82.49     & 2× \\
PartPrompt     & 46.06  & 37.83   &  10.44   & 20.00   & 30.18      & 82.57     & 2× \\
LLMLingua-2       & 45.99  & 38.77   & 11.18   & 20.42   & 30.81      & 82.60     & 2× \\
$\text{SCOPE}_N$      & \textbf{46.51}  & 38.97   & 11.51   & 20.53   & \textbf{30.95}      & \textbf{82.69}     & 2× \\
$\text{SCOPE}_D$      & 46.38  & \textbf{39.03}   & \textbf{11.59}   & \textbf{20.56}   & \textbf{30.95}      & \textbf{82.69}     & 2× \\
\cmidrule(lr){1-1}\cmidrule(lr){2-7}\cmidrule(lr){8-8}
Original            & 47.29   & 40.22 & 12.62 & 21.46 & 31.93 & 83.13 & 1× \\
\bottomrule
\end{tabular}
}
\label{tab:pubmed_summarization}
\end{table*}

\begin{table*}[t]
\centering
\scriptsize
\setlength{\tabcolsep}{2.5pt}
\caption{Ablation study for different embedding models on GovReport-Summarization dataset and different summarization models on Arxiv-Summarization dataset, across different compression ratios on GPT-4o-mini.}
\resizebox{\textwidth}{!}{
\begin{tabular}{l|cccccc|r}
\toprule
\multicolumn{8}{c}{\textbf{GovReport-Summarization over different embedding models}} \\
\midrule
Method            & BLEU (\%) & ROUGE-1 (\%) & ROUGE-2 (\%) & ROUGE-L (\%) & ROUGE-Lsum (\%) & BERTScore (\%) & $\rho$ \\
\cmidrule(lr){1-1}\cmidrule(lr){2-7}\cmidrule(lr){8-8}
$\text{SCOPE}_N$ W/T all-MiniLM   & \textbf{60.52} & 43.59 & \textbf{14.00} & 19.20 & 24.76 & \textbf{81.86} & 5× \\
$\text{SCOPE}_N$ W/T Jina-Embedding         & 60.20 & \textbf{43.66} & 13.85 & \textbf{19.26} & \textbf{24.79} & 81.75 & 5× \\
Difference                 & 0.32 & 0.07 & 0.15 & 0.06 & 0.03 & 0.11 & 5× \\
\cmidrule(lr){1-1}\cmidrule(lr){2-7}\cmidrule(lr){8-8}
$\text{SCOPE}_N$ W/T all-MiniLM   & \textbf{60.86} & \textbf{43.69} & \textbf{14.10} & 19.36 & \textbf{24.85} & \textbf{81.97} & 3× \\
$\text{SCOPE}_N$ W/T Jina-Embedding         & 60.58 & 43.61 & 14.05 & \textbf{19.44} & 24.83 & 81.94 & 3× \\
Difference                 & 0.28 & 0.08 & 0.05 & 0.08 & 0.02 & 0.03 & 3× \\
\cmidrule(lr){1-1}\cmidrule(lr){2-7}\cmidrule(lr){8-8}
$\text{SCOPE}_N$ W/T all-MiniLM   & \textbf{61.35} & \textbf{43.79} & \textbf{14.44} & \textbf{19.58} & \textbf{24.95} & \textbf{82.17} & 2× \\
$\text{SCOPE}_N$ W/T Jina-Embedding         & 60.95 & 43.57 & 14.10 & 19.51 & 24.88 & 82.05 & 2× \\
Difference                 & 0.40 & 0.22 & 0.34 & 0.07 & 0.07 & 0.12 & 2× \\
\toprule
\multicolumn{8}{c}{\textbf{Arxiv-Summarization over different summarization models}} \\
\midrule
Method            & BLEU (\%) & ROUGE-1 (\%) & ROUGE-2 (\%) & ROUGE-L (\%) & ROUGE-Lsum (\%) & BERTScore (\%) & $\rho$ \\
\cmidrule(lr){1-1}\cmidrule(lr){2-7}\cmidrule(lr){8-8}
$\text{SCOPE}_N$ W/T BART              & \textbf{38.09} & \textbf{37.75} & 10.31 & \textbf{20.16} & \textbf{29.12} & \textbf{80.93} & 5× \\
$\text{SCOPE}_N$ W/T Google Pegasus    & 36.36 & 37.28 & \textbf{10.38} & 19.95 & 28.62 & 80.72 & 5× \\
Difference      & 1.73 & 0.47 & 0.07 & 0.21 & 0.50 & 0.21 & 5× \\
\cmidrule(lr){1-1}\cmidrule(lr){2-7}\cmidrule(lr){8-8}
$\text{SCOPE}_N$ W/T BART              & \textbf{38.26} & \textbf{38.17} & 10.55 & 20.24 & \textbf{29.40} & \textbf{80.98} & 3× \\
$\text{SCOPE}_N$ W/T Google Pegasus    & 36.86 & 38.03 & \textbf{11.05} & \textbf{20.40} & 29.18 & 80.96 & 3× \\
Difference      & 1.40 & 0.14 & 0.50 & 0.16 & 0.22 & 0.02 & 3× \\
\cmidrule(lr){1-1}\cmidrule(lr){2-7}\cmidrule(lr){8-8}
$\text{SCOPE}_N$ W/T BART              & \textbf{38.14} & 38.66 & 11.04 & 20.67 & 29.78 & 81.09 & 2× \\
$\text{SCOPE}_N$ W/T Google Pegasus    & 37.22 & \textbf{39.08} & \textbf{11.87} & \textbf{21.03} & \textbf{30.03} & \textbf{81.15} & 2× \\
Difference      & 0.92 & 0.42 & 0.83 & 0.36 & 0.25 & 0.06 & 2× \\
\bottomrule
\end{tabular}
}
\label{tab:embedding_summ_horizontal_strict}
\end{table*}

\subsection{Additional experiments}\label{sec:appendix-addi-exp}

\subsubsection{Impact of different embedding models}

We evaluate SCOPE with different embedding models on GovReport-Summarization dataset under all 3 compression ratios. The embedding model in \ours\ for all other experiments is \textit{All-MiniLM} (\cite{allminilm}), while the competitor model in this experiment is \textit{Jina-embedding} from JinaAI (\cite{gunther2023jina}).
In Table ~\ref{tab:embedding_summ_horizontal_strict}, we report the scores and absolute score differences between \ours\ using different embedding models. The two embedding models result in a negligible performance difference, where the largest absolute difference is 0.40\%, and lowest is 0.02\%. This shows a stable performance of SCOPE when switching embedding model, indicating the outstanding adaptivity of SCOPE.

\subsubsection{Impact of different summarization models}

We evaluate SCOPE with different summarization models on Arxiv-Summarization dataset under all 3 compression ratios. The summarization model in \ours\ for all other experiments is \textit{BART} (\cite{lewis2019bart}), while the competitor model in this experiment is \textit{Pegasus} from Google (\cite{zhang2020pegasus}).
In Table ~\ref{tab:embedding_summ_horizontal_strict}, we report the scores and absolute score differences between \ours\ using different summarization models. The two summarization models result in a negligible performance difference, where the largest absolute difference is 1.73\%, and lowest is 0.02\%. This shows a stable performance of SCOPE when switching across different summarization models, indicating the outstanding adaptivity of SCOPE.

\subsubsection{Evaluation on Multi-Document Summarization}
\label{sec:multinews}

To further evaluate SCOPE on multi-document inputs, we conduct additional experiments on the Multi-News dataset from LongBench. Unlike the single-document summarization datasets used in the main evaluation, Multi-News requires summarizing information collected from multiple news documents. Table~\ref{tab:multinews} reports the results under 5$\times$, 3$\times$, and 2$\times$ compression ratios. SCOPE consistently achieves the highest scores across all evaluated metrics and compression ratios. For example, under 5$\times$ compression, SCOPE achieves ROUGE-1 and ROUGE-2 scores of 36.84 and 9.65, compared with 34.80 and 7.71 from LLMLingua-2. These results show that SCOPE remains effective for multi-document summarization.

\begin{table*}[t]
\centering
\scriptsize
\setlength{\tabcolsep}{4pt}
\caption{Evaluation on the LongBench Multi-News dataset at different compression ratios on GPT-4o-mini.}
\resizebox{\textwidth}{!}{
\begin{tabular}{l|cccccc|r}
\toprule
\multicolumn{8}{c}{\textbf{Multi-News}} \\
\midrule
Method            & BLEU (\%) & ROUGE-1 (\%) & ROUGE-2 (\%) & ROUGE-L (\%) & ROUGE-Lsum (\%) & BERTScore (\%) & $\rho$ \\
\cmidrule(lr){1-1}\cmidrule(lr){2-7}\cmidrule(lr){8-8}

Selective-Context & 40.40 & 32.97 & 5.91 & 15.11 & 18.57 & 77.78 & 5× \\
LLMLingua         & 37.67 & 29.29 & 4.06 & 13.91 & 16.90 & 76.11 & 5× \\
LongLLMLingua     & 39.15 & 31.68 & 5.36 & 14.97 & 17.71 & 77.21 & 5× \\
PartPrompt        & 39.94 & 33.18 & 5.98 & 15.48 & 18.74 & 77.65 & 5× \\
LLMLingua-2       & 43.76 & 34.80 & 7.71 & 16.06 & 19.00 & 78.71 & 5× \\
SCOPE             & \textbf{45.54} & \textbf{36.84} & \textbf{9.65} & \textbf{17.57} & \textbf{20.39} & \textbf{79.48} & 5× \\

\cmidrule(lr){1-1}\cmidrule(lr){2-7}\cmidrule(lr){8-8}

Selective-Context & 41.92 & 34.31 & 7.43 & 16.09 & 19.11 & 78.59 & 3× \\
LLMLingua         & 38.95 & 31.20 & 4.93 & 14.89 & 17.77 & 77.29 & 3× \\
LongLLMLingua     & 40.59 & 33.32 & 6.23 & 15.54 & 18.66 & 77.84 & 3× \\
PartPrompt        & 41.28 & 34.67 & 7.19 & 16.44 & 19.67 & 78.40 & 3× \\
LLMLingua-2       & 45.12 & 36.22 & 8.83 & 17.03 & 19.93 & 79.13 & 3× \\
SCOPE             & \textbf{45.88} & \textbf{37.24} & \textbf{10.12} & \textbf{17.71} & \textbf{20.47} & \textbf{79.60} & 3× \\

\cmidrule(lr){1-1}\cmidrule(lr){2-7}\cmidrule(lr){8-8}

Selective-Context & 43.49 & 35.47 & 8.51 & 16.76 & 19.81 & 79.08 & 2× \\
LLMLingua         & 40.63 & 32.94 & 6.25 & 15.58 & 18.52 & 78.01 & 2× \\
LongLLMLingua     & 41.99 & 34.28 & 7.25 & 16.26 & 19.30 & 78.55 & 2× \\
PartPrompt        & 42.37 & 35.76 & 8.34 & 16.98 & 20.30 & 78.80 & 2× \\
LLMLingua-2       & 45.29 & 36.46 & 9.35 & 17.31 & 20.01 & 79.20 & 2× \\
SCOPE             & \textbf{46.54} & \textbf{37.77} & \textbf{10.29} & \textbf{17.95} & \textbf{20.87} & \textbf{79.81} & 2× \\

\bottomrule
\end{tabular}
}
\label{tab:multinews}
\end{table*}

\subsubsection{Evaluation on the Full Arxiv-Summarization Test Set}
\label{sec:full_arxiv}

The main experiments use up to 2,000 randomly sampled examples from each dataset due to the computational cost of evaluating multiple compression methods followed by downstream LLM inference. To examine whether this sampling affects the experimental conclusions, we additionally evaluate all methods on the full Arxiv-Summarization test set containing 6,440 examples. Table~\ref{tab:full_arxiv} reports the results under 5$\times$ compression. The relative ranking of the evaluated methods remains unchanged, with SCOPE achieving the highest scores across all metrics. This result suggests that the conclusions obtained from the sampled evaluation are consistent with those on the full test set.

\begin{table*}[t]
\centering
\scriptsize
\setlength{\tabcolsep}{4pt}
\caption{Evaluation on the full Arxiv-Summarization test set (6,440 examples) at 5$\times$ compression on GPT-4o-mini.}
\resizebox{\textwidth}{!}{
\begin{tabular}{l|cccccc|r}
\toprule
\multicolumn{8}{c}{\textbf{Full Arxiv-Summarization Test Set}} \\
\midrule
Method            & BLEU (\%) & ROUGE-1 (\%) & ROUGE-2 (\%) & ROUGE-L (\%) & ROUGE-Lsum (\%) & BERTScore (\%) & $\rho$ \\
\cmidrule(lr){1-1}\cmidrule(lr){2-7}\cmidrule(lr){8-8}

LLMLingua         & 31.82 & 31.55 & 5.23  & 17.10 & 24.92 & 78.82 & 5× \\
LongLLMLingua     & 32.68 & 33.45 & 6.67  & 17.81 & 26.03 & 79.40 & 5× \\
PartPrompt        & 32.81 & 33.81 & 7.17  & 18.13 & 26.47 & 79.45 & 5× \\
LLMLingua-2       & 34.43 & 35.87 & 8.38  & 18.78 & 27.47 & 80.17 & 5× \\
SCOPE             & \textbf{37.05} & \textbf{38.88} & \textbf{11.66} & \textbf{20.83} & \textbf{29.76} & \textbf{81.03} & 5× \\

\bottomrule
\end{tabular}
}
\label{tab:full_arxiv}
\end{table*}

\subsubsection{Impact of Different Compression Model Backbones}
\label{sec:compression_backbone}
 
To further examine whether SCOPE depends on a particular summarization backbone, we replace BART with Llama-3-8B-Instruct as the compression model and evaluate both models on Longbench GovReport and Multi-News. Table~\ref{tab:compression_backbone} reports the results under 5$\times$, 3$\times$, and 2$\times$ compression ratios. The two backbones exhibit similar performance across both datasets and all compression ratios. Although BART achieves slightly higher scores in most settings, replacing it with Llama-3-8B-Instruct preserves the overall performance trend, indicating that SCOPE is not limited to a specific compression model backbone.

\begin{table*}[t]
\centering
\scriptsize
\setlength{\tabcolsep}{6pt}
\caption{Compression model backbone replacement study using BART and Llama-3-8B-Instruct on Longbench GovReport and Multi-News.}
\resizebox{0.75\textwidth}{!}{
\begin{tabular}{l|r|l|ccc}
\toprule
Dataset & $\rho$ & Compression Model & ROUGE-1 (\%) & ROUGE-2 (\%) & BERTScore (\%) \\
\midrule

GovReport & 5× & BART        & \textbf{27.99} & \textbf{10.89} & \textbf{82.73} \\
          & 5× & Llama-3-8B & 27.40 & 10.56 & 81.68 \\
\cmidrule(lr){2-6}
          & 3× & BART        & \textbf{28.17} & \textbf{11.43} & \textbf{82.78} \\
          & 3× & Llama-3-8B & 27.83 & 11.12 & 81.89 \\
\cmidrule(lr){2-6}
          & 2× & BART        & \textbf{28.32} & \textbf{11.54} & \textbf{82.90} \\
          & 2× & Llama-3-8B & 27.84 & 11.38 & 82.03 \\

\midrule

Multi-News & 5× & BART        & \textbf{36.84} & 9.65          & \textbf{79.48} \\
           & 5× & Llama-3-8B & 36.61          & \textbf{9.74} & 79.36 \\
\cmidrule(lr){2-6}
           & 3× & BART        & \textbf{37.24} & \textbf{10.12} & \textbf{79.60} \\
           & 3× & Llama-3-8B & 36.68 & 9.94 & 79.32 \\
\cmidrule(lr){2-6}
           & 2× & BART        & \textbf{37.77} & \textbf{10.29} & \textbf{79.81} \\
           & 2× & Llama-3-8B & 36.88 & 9.95 & 79.53 \\

\bottomrule
\end{tabular}
}
\label{tab:compression_backbone}
\end{table*}

\end{document}